\documentclass{article}

\newcommand{\define}{:=}
\newcommand{\ud}{\mathsf{d}}
\newcommand{\ellone}{\ud_1}
\newcommand{\ellrelinf}{\ud_{\infty}}
\newcommand{\RD}{\mathsf{RD}}

\newcommand{\resource}{N_{\mbox{\tiny$\Sigma$}}}
\newcommand{\population}{P_{\mbox{\tiny$\Sigma$}}}

\usepackage{xcolor}
\usepackage{xspace,enumerate}
\usepackage{subcaption}
\usepackage{arxiv}
\usepackage[utf8]{inputenc} 
\usepackage[T1]{fontenc}    
\usepackage{hyperref}       
\usepackage{url}            
\usepackage{booktabs}       
\usepackage{amsfonts}       
\usepackage{nicefrac}       
\usepackage{microtype}      
\usepackage{lipsum}
\usepackage{graphicx}
\graphicspath{ {./images/} }
\usepackage{mathtools}
\usepackage{amsmath}
\usepackage{upgreek}
\usepackage{multirow}
\definecolor{red}{rgb}{1,0,0}
\definecolor{green}{rgb}{0,0.6,0}
\definecolor{blue}{rgb}{0,0,0.8}

\newtheorem{theorem}{Theorem}[section]

\newtheorem{proposition}[theorem]{Proposition}

\def\FullBox{\hbox{\vrule width 6pt height 6pt depth 0pt}}

\def\qed{\ifmmode\qquad\FullBox\else{\unskip\nobreak\hfil
\penalty50\hskip1em\null\nobreak\hfil\FullBox
\parfillskip=0pt\finalhyphendemerits=0\endgraf}\fi}

\def\qedsketch{\ifmmode\Box\else{\unskip\nobreak\hfil
\penalty50\hskip1em\null\nobreak\hfil$\Box$
\parfillskip=0pt\finalhyphendemerits=0\endgraf}\fi}

\def\to{\rightarrow}

\def\epsilon{\varepsilon}

\def\phi{\varphi}

\renewcommand{\bar}{\overline}

\newcommand{\Psymb}{\mathbb{P}}

\DeclareMathOperator*{\ProbOp}{\Psymb}

\renewcommand{\Pr}{\ProbOp}

\newcommand{\prob}[1]{\Pr\left({#1}\right)}

\newcommand{\E}[1]{\ExpOp\left[{#1}\right]}

\usepackage{nicefrac}

\newfont{\inhead}{eufm10 scaled\magstep1}

\newcommand{\ue}{\mathrm{e}}

\DeclareMathOperator*{\argmin}{arg\,min}

\title{Modeling Access Differences to Reduce Disparity in Resource Allocation}

\author{
 Kenya S. Andrews \\
  University of Illinois at Chicago\\
  \texttt{kandre32@uic.edu} \\
   \And
 Mesrob I. Ohannessian \\
  University of Illinois at Chicago \\
  \texttt{mesrob@uic.edu} \\
  \And
 Tanya Berger-Wolf \\
  Ohio State University\\
  \texttt{berger-wolf.1@osu.edu} \\
}

\begin{document}
\maketitle
\begin{abstract}
Motivated by COVID-19 vaccine allocation, where vulnerable subpopulations are simultaneously more impacted in terms of health and more disadvantaged in terms of access to the vaccine, we formalize and study the problem of resource allocation when there are inherent access differences that correlate with advantage and disadvantage. We identify reducing resource disparity as a key goal in this context and show its role as a proxy to more nuanced downstream impacts. We develop a concrete access model that helps quantify how a given allocation translates to resource flow for the advantaged vs. the disadvantaged, based on the access gap between them. We then provide a methodology for access-aware allocation. Intuitively, the resulting allocation leverages more vaccines in locations with higher vulnerable populations to mitigate the access gap and reduce overall disparity. Surprisingly, knowledge of the access gap is often not needed to perform access-aware allocation. To support this formalism, we provide empirical evidence for our access model and show that access-aware allocation can significantly reduce resource disparity and thus improve downstream outcomes. We demonstrate this at various scales, including at county, state, national, and global levels.
\end{abstract}


\section{Introduction}

All common resources must be allocated in some way to members of a society. During such an allocation, the goal should be to optimize the benefits of the resources for that society. To be transformational, the way we allocate needs to be not only superficially fair but also deeply just and willfully acting to mitigate unfairness. We learned a great deal about what happens when we are not intentional in our allocation efforts through the COVID-19 virus health crisis. Yet, amidst the rise in monkeypox cases, we find ourselves in familiar territory with shortages and allocation issues for the monkeypox virus vaccines \cite{monkeypox_late,monkeypox_distribution}. For most of the COVID-19 pandemic, government officials and healthcare workers have been rapidly seeking ways to minimize loss caused by the COVID-19 virus. In this search, they found the virus affects members of the vulnerable population more severely; namely people with underlying conditions, members of racial minority groups, those who lack personal transportation, those of lower economic status, and older people \cite{ncrid}. Vulnerable people experience worse short and long term health side effects and higher mortality rates \cite{cdc_mortality_2022}. Additionally, the impact on cities and states when members of the vulnerable populations get sick from COVID-19 leads to higher hospitalization rates, lack of healthcare staff availability, and less hospital space to care for other conditions. Though these realities were recognized, many of the plans that were developed to minimize loss, including vaccination, did not fully take this into consideration. Instead, when vaccines began to be distributed, they were distributed according to age (oldest) or career (most essential workers) \cite{jean2021vaccine}. Additionally, points of distribution for vaccines were receiving proportional amounts of vaccines \cite{covid_distribution_plan}. It is seemingly ideal to have such an allocation, but there are several examples where this logic breaks. For example, age is not the only good measure because we see Black Americans have a shorter life expectancy than their White counterparts due to health inequities \cite{kirby2010unhealthy}. Thus, only focusing on age exacerbates these issues for Black Americans. 

Motivated by COVID-19 vaccine allocation, we formalize the problem of fair resource allocation when there are inherent access differences that correlate with advantage and disadvantage. We make and support three key premises:

\begin{itemize}
    \item[(A)] The population is stratified into disadvantaged and advantaged subpopulations.
    \item[(B)] The disadvantaged subpopulation is more vulnerable to the lack of resources.
    \item[(C)] The disadvantaged subpopulation has obstacles in acquiring allocated resources.
\end{itemize}

In Sec. \ref{sec:problem-description}, we describe the formal problem by closely following premises (A) and (B). Our first contribution there is to highlight reducing \textbf{resource disparity}  between advantaged and disadvantaged subpopulations as a key goal of resource allocation. Then, in Sec. \ref{sec:access}, our second contribution is a concrete \textbf{access model}, which we develop through premise (C). This allows us to determine how resources would flow to the disadvantaged vs. advantaged subpopulations, upon performing a given allocation. In Sec. \ref{sec:impact}, we show that resource disparity can be effectively thought of as a proxy to more complex impact disparities downstream. In Sec. \ref{sec:allocation}, we combine the formal problem with the access model to obtain our third contribution, an \textbf{access-aware allocation} algorithm that reduces resource disparity while maintaining approximate geographic proportionality. A key finding here is that access-aware allocation can often be performed without explicit knowledge of the access gap between the subpopulations. We acknowledge that there are many contributors to access disparities other than disadvantage. That said, in Sec. \ref{sec:access-evidence} we give empirical evidence in support of our access model. Then, in Sec. \ref{sec:behavior-outcome}, we revisit the COVID-19 vaccine allocation problem as a use case. Our fourth contribution is, we show \textbf{significant reduction} in resource disparity and thus downstream impact can be achieved with minimal deviation from proportionality. In Sec. \ref{sec:discussions}, we conclude the paper with discussion points. We start with a survey of related work in Sec. \ref{sec:related-work}.

\section{Related Work} \label{sec:related-work}

The importance of considering equity and justice when allocating resources has not been lost on the scientific community, especially for the COVID-19 vaccine distribution problem, the primary case study here. The role of access differences and its interplay with vulnerability and disadvantage is often alluded to, but not directly formalized as we do.

For example, in \cite{jean2021vaccine}, the authors note that those 65 and older account for 80\% of deaths from COVID-19; however, they warn that ``if this is done without explicit attention to promoting health equity, it will, once again, exacerbate major health disparities.’’ They attribute this to the fact that: making a vaccination appointment requires substantial time, reliable internet access and technology, ability to travel to a vaccination site, and trust in the safety of the vaccine. Note that inability to do any of these and the existence of pre-exisiting health conditions correlates with a person being more vulnerable. Several solutions are offered in \cite{jean2021vaccine}, including prioritizing hardest hit and lowest economic ZIP Codes, which is very much in line with the access-aware allocation solutions that come out of our explicit model. Other suggestions are: local vaccine education, decreasing transportation barriers, and simplifying registration procedures. Though we do not discuss these, we do acknowledge that they could aid in grassroots efforts of vaccine administration.

In \cite{wedlund2021new}, the authors develop a tool to identify people who will benefit most from receiving the vaccine at a specific time period. They argue that those at most risk of contracting should be prioritized.  However, access differences are not made explicit, and this focus does not necessarily address minimizing the harshest effects of contracting COVID-19.

In \cite{estiri2021predicting}, data from electronic hospital records (EHRs) is used to train a model which can determine the likelihood of a person to die from contracting COVID-19. While we do not depend on health records to determine whether a person would die from COVID-19, we do use the CDC’s full definition of what it means to be vulnerable. The findings of \cite{estiri2021predicting} (see also \cite{garg2020hospitalization} for early variant findings) clearly show affect the level of hospitalization and death risk. It is also worth noting that health records for racial minorities, lower-income people, and those who are in more rural areas tend to not be complete profiles due to access barriers \cite{norris2008race}.

Many data-driven methods to guide vaccine allocation have also been proposed. In \cite{davahli2021optimizing}, the authors cluster states with similarly reported effective reproduction numbers\footnote{an epidemiological property that describes the expected number of new cases due directly to one case} to look for an optimal distribution. Their focus does not include equity and is more on preventing spread, which we do not address. By concentrating on resource equity between vulnerable and non-vulnerable subpopulations, we focus on who would benefit most from receiving enough vaccines.

In \cite{chen2020allocation}, allocation under limited resources is addressed but without considerations of equity, disparity, or access. There, the importance of dynamic allocation is emphasized, which we do not address. In \cite{perakis2021covid}, the authors consider overall COVID-19 impacts (death, job loss, and changes in the economy). They build a linear optimization method to predict testing allocation, based on predicting the influx of cases, over different time periods to maximize detection of the virus amongst different population groups (symptomatic, asymptomatic, and contact tracing). They incorporate a basic geographic notion of equity, by allocating a minimum number of resources at each location, but do not model access disparity. In a different relevant line of work, resources are allocated to increase direct (spatial) access. For example, in \cite{zhang_yang_pan_2021}, the placement of hospital beds is optimized using a model of spatial accessibility. This does not involve an explicit relationship between vulnerability and resource acquisition but is relevant nonetheless in that it centers on issues of access by the disadvantaged and findings include resulting improvements in equity also.

In the theoretical CS and OR communities, models of equity and allocation have also been proposed. Recently, \cite{donahue2020fairness} studied resource allocation across groups with varying demand distributions and showed tradeoffs between utilization and fairness across groups. The varying demand distributions could be interpreted as access differences, but it is important to note that upon instantiation of the demand, there is no inherent difference in the ease with which the resource is acquired. More importantly, in that model, individual allocations can be made for each group, which may not ideally describe shared public resource allocation, save for assistance models that occur through an application process. Though we focus on access differences in physical resource acquisition, we acknowledge the importance of differences in access to information and their impact, e.g., as is done in \cite{garg2020dropping} for standardized testing. Of course, there are many dimensions beyond access, such as dynamics and correlations, that further complicate resource acquisition. For example, this was explored in depth by \cite{manshadi2021fair} for rationing medical supplies in the COVID-19 backdrop.

A key theme in the current paper is that the disadvantaged lack ease of access to resources and, simultaneously, are disproportionally affected by the burden of lack of resource \cite{ centers_for_disease_control_and_prevention_2021, ncdhhs}. Sometimes they can be targeted directly (e.g., mobile clinics, government-supplied PPE, etc). Though these are helpful solutions, these efforts are expensive and resources often run out before having had sufficient impact \cite{simmons-duffin_2022}. Policies that place more of these resources where they are needed, such as the access-aware allocation methods proposed here, can boost resource acquisition indirectly. Reaching vulnerable populations and mitigating the obstacles they face are critical in making optimization and data-driven methods achieve their true clinical potential \cite{zhou2021clinical, marwaha_kvedar_2022}. We believe explicitly modeling access differences is an essential tool for achieving this potential.

\section{Problem Description} \label{sec:problem-description}

\paragraph{Allocation Task and Notation} We are tasked with allocating an amount $\resource$ of resources, assumed to be arbitrarily divisible, across $k$ locations indexed by $j \in [k] \define \{1,\cdots,k\}$. The decision variables are the amounts to be allocated to each location $j$, denoted by $N_j$, or equivalently the fraction of the whole that these represent, denoted by $n_j \define N_j/\resource$. It is desirable to entirely allocate the resources, so that $\sum_j N_j = \resource$ and $\sum_j n_j = 1$. We write $\mathbf{n}\define(n_j)_{j\in[k]}$. The implicit target for this allocation are the population sizes $P_j$ at each location. Using parallel notation to $\resource$, let $\population=\sum_j P_j$ be the total population size and $p_j \define P_j/\population$ be the fraction of the population at location $j$, so that $\sum_j p_j=1$. We write $\mathbf{p}\define(p_j)_{j\in[k]}$. Resources are assumed to be limited, $\resource<\population$, and we parametrize this explicitly as $\alpha \define \resource/\population \in (0,1)$.

\paragraph{Disadvantage vs. Advantage} A central premise of this work, Premise (A), is that the population is stratified by disadvantage and that this stratification varies across locations. The \emph{disadvantaged} should be regarded as particularly vulnerable in the absence of a resource. We adopt the simplest model that encapsulates this, by using a parameter $\beta_j \in [0,1]$, which describes the fraction at location $j$ who is a particularly disadvantaged subpopulation. While the remainder $1-\beta_j$ of the population may also have some vulnerability, we will describe them as \emph{advantaged}. This stratification influences the allocation task through a difference in how allocated resources are acquired by each subpopulation. At any given location $j$, given $N_j$ resources, we assume there exists an acquisition function $\rho$, that determines the fraction of $N_j$ acquired by the disadvantaged, while the remainder $1-\rho$ is acquired by the advantaged.

How can we characterize $\rho$? We can only do this through a clear model of access that describes the process by which allocated resources are acquired by each subpopulation and the difference between them. This is the motivation of Sec. \ref{sec:access}, which elaborates our proposed access model theoretically --- this is later supported empirically in Sec. \ref{sec:access-evidence}. The resulting $\rho$ is found to depend on the local quantities $N_j$, $P_j$, and $\beta_j$, as well as a global parameter $\eta$ that describes the \emph{access gap} between disadvantaged and advantaged subpopulations. In what follows, for conciseness, we refer to the value that $\rho$ takes at location $j$ as $\rho_j$. For reference, please refer to Table \ref{tab:glossary} for a glossary of the main notations.

\begin{table}[b]
    \footnotesize
    \centering
    \begin{tabular}{r|lcr|l}
         $\population$ & total population size  & \hspace{1pt} &  $\mathbf{n}$ & vector of all $n_j$\\
         $\resource$  &  total resources available  & \hspace{1pt} &  $\beta_j$ & percentage of disadvantaged at location $j$\\
         $\alpha$ &  resource availability, $\resource/\population$  & \hspace{1pt} &  $\rho_j$ & percentage of $N_j$ acquired by the disadvantaged\\
         $k$ & number of locations   & \hspace{1pt} &  $\eta$ & access gap between disadvantaged and advantaged \\
         $P_j$ & population size at location $j\in\{1,\cdots,k\}$    & \hspace{1pt} &  $\ud_1$ & $\ell_1$ distance \\
         $N_j$ & resources allocated to location $j$    & \hspace{1pt} &   $\ud_\infty$ & relative $\ell_\infty$ distance \\
         $p_j$ & fraction of total population at location $j$, $P_j / \population$    & \hspace{1pt} &  $\varepsilon$ & tolerance in distance from proportionality \\
         $n_j$ & fraction of total resources allocated to $j$, $N_j / \resource$   & \hspace{1pt} &  $\mathbf{n}^\star_\cdot$ & optimal allocation when distance $\ud_\cdot$ is used \\
         $\mathbf{p}$ & vector of all $p_j$   & \hspace{1pt} &  $\RD$ & resource disparity, Eq. \eqref{eq:disparity} 
    \end{tabular}
    \caption{Glossary of main notations}
    \label{tab:glossary}

\end{table}

\paragraph{Resource Rate Disparity} Upon an allocation $\mathbf{n}$, each subpopulation acquires part of the $\resource$ resources. The disadvantaged acquire $\sum_j \rho_j N_j = \resource \sum_j \rho_j n_j$, while the advantaged acquire the rest $\sum_j (1-\rho_j) N_j = \resource \sum_j (1-\rho_j) n_j$. As a result, resource acquisition rate may vary across the two subpopulations, which we quantify through \emph{rate disparity}:
\begin{equation} 
    \RD = \frac{\sum_j (1-\rho_j) N_j}{\sum_j (1-\beta_j) P_j} - \frac{\sum_j \rho_j N_j}{\sum_j \beta_j P_j} = \alpha \frac{\sum_j (1-\rho_j) n_j}{\sum_j (1-\beta_j) p_j} - \alpha \frac{\sum_j \rho_j n_j}{\sum_j \beta_j p_j}.\label{eq:disparity}
\end{equation}

If the notion of disadvantage is true to its name, then $\RD$ will typically be non-negative, as the resource acquisition rate will be higher within the advantaged group. As we see in Sec. \ref{sec:empiricalval}, this is indeed the case for the vaccine allocation problem. Moreover, it is desirable not only to have \textbf{$\RD$ close to zero}, but also to make it \textbf{negative}, as downstream outcomes may improve with higher allocation rates among the disadvantaged. We expand on this in Sec. \ref{sec:impact}. This codifies Premise (B) and is the reason why the definition in Eq. \eqref{eq:disparity} does not enforce positivity. That said, $|\RD|$ may also be handled with minimal modifications within the present methodology.

\paragraph{Proportional Allocation and Geographic Disparity} \looseness=-1 The baseline allocation is proportional to the population, that is $n_j = p_j$, for all $j$. This is fundamentally a notion of geographic equity that is an expected and desirable property of a potential allocation in many policies, including some recent resource allocation work, e.g. in COVID-19 testing allocation \cite{perakis2021covid}. To allow ourselves to remain close to this property, we relax the strict equality of $n_j=p_j$ for all $j$ to a distance on the $k$-simplex between $\mathbf{n}$ and $\mathbf{p}$. One can think of this as a form of headroom for social intervention, up to a limit that could be potentially determined by law or by agreement. We primarily consider the absolute $\ell_1$ and relative $\ell_\infty$ distances:
\begin{equation}
    \ellone(\mathbf{n},\mathbf{p}) = \sum_j |n_j-p_j| \qquad\qquad
    \ellrelinf(\mathbf{n},\mathbf{p}) = \max_j \left|\frac{n_j}{p_j}-1\right| \label{eq:distances}
\end{equation}
Note that $\ellrelinf$ is the more stringent of the two distances as $\ellone = \sum_j p_j \left|\frac{n_j}{p_j}-1\right| \leq \ellrelinf$. (The methodology could also handle absolute $\ell_\infty$ and relative $\ell_1$ distances, given by $\max_j |n_j-p_j|$ and $\sum_j \left|\frac{n_j}{p_j}-1\right|$ respectively, and weighted variants.)

\paragraph{Allocation Objective}

It is worth remarking that if $\rho_j = \beta_j$, that is if disadvantaged and advantaged subpopulations get their proportional shares of the allocated resources at each location, then proportional allocation yields zero rate disparity, $\RD=0$. Intuitively, this is a scenario where there is no access gap between the disadvantaged and advantaged. When there is a gap, we expect proportional allocation to not yield the best rate disparity. How can we allocate resources to improve disparity across subpopulations while maintaining a reasonable geographic parity? This trade-off leads to the central objective of this paper, which we can express at a high level as:
\begin{equation}
 \begin{array}{lcrcl}
    \mathbf{n}^\star_{\cdot} & = & \argmin &\;& 
    \RD(\mathbf{n},\mathbf{p}, \resource, \population, \beta,\eta) \\
       & & \textrm{subject to} &\;& \ud_{\cdot}(\mathbf{n},\mathbf{p})\leq \varepsilon\\
        & & &\;& \alpha \mathbf{n} \leq \mathbf{p},
    \end{array} \label{eq:abstract-objective}
\end{equation}
where the last inequality is a \emph{no-waste constraint}, which prohibits the allocation of more resources than people at an location $j$: $N_j \leq P_j$. The $\cdot$ can be either $1$ or $\infty$, indicating the distance used.


\section{Modeling Access} \label{sec:access}

In this section, our scope is a single location $j$. As such, for clarity, we drop all $j$ subscripts.


\subsection{Resource Acquisition}

To model access differences across disadvantaged and advantaged populations, we need to model how resources are acquired upon allocation. For this, we adopt a simple yet effective model: \emph{Poisson acquisition until saturation or exhaustion}. Each subpopulation acquires a unit of resource according to an independent Poisson arrival process, which stops when one of two things happen: (1) an amount of resource equal to the subpopulation size is acquired by the subpopulation (saturation) or (2) a total amount of resource equal to the allocated size is acquired (exhaustion). This model allows us to account for access differences via the relative rate of each Poisson process. In what follows, we focus on a specific location $j$ with population $P$ where $\beta$ fraction of the population is disadvantaged and $N$ resources have been already allocated. Our goals are to specify the acquisition model, describe access gaps, and derive the shape of the acquisition function $\rho$. We assume that resources and subpopulation sizes are integers.

\paragraph{Base Model} Let $\lambda^-$ and $\lambda^+$ be the rates of acquisition of the disadvantaged and advantaged respectively. When there are no access gaps, these rates reflect only the relative sizes of the subpopulations. We set the expected time for a single resource acquisition per person in this case as our unit of time, and treat the two populations as a split. This gives us the base rates of $\lambda^- = \beta P$ and $\lambda^+ = (1-\beta) P$.

\paragraph{Access Gap} Premise (C) states that the disadvantaged face obstacles in acquiring resources. We can model this as a \emph{slowdown} relative to the base rate. Let $\eta$ be the (relative) access gap between the disadvantaged and advantaged populations, where $\eta=1$ signifies no gap and $\eta=0$ is the limit of infinite gap. Under this model, the modified rate for the disadvantaged becomes $\lambda^- = \eta \beta P$ while the rate of the advantaged remains unchanged $\lambda^+ = (1-\beta) P$.


\subsection{Na\"ive Acquisition Function}

With this model, we can now concretely define the acquisition function $\rho$ to be the expected fraction of the total allocated $N$ resources that will be acquired by the disadvantaged. To develop intuition, let us start with a na\"ive characterization that ignores saturation. This can be thought of as the limit of very scarce resources ($\alpha\to 0$), so much so that no subpopulation in any location can be saturated. In this case, subpopulations acquire resources until exhaustion. If $\ddot U_t$ represents the disadvantaged process and $\ddot V_t$ represents the advantaged process, then the exhaustion time can be defined as $\ddot S=\min \{ t~:~ \ddot U_t+ \ddot V_t\geq N\}$. With this, the acquisition function can be defined as $\ddot \rho \define \E{\ddot U_{\ddot S}} / N$. The $\ddot{\phantom{\rho}}$ notation is used to evoke \emph{``na\"ive''}, and not second derivative. We have (the proof is given in the supplementary material):

\begin{proposition} \label{prop:naive-rho}
    If saturation is ignored, the na\"ive acquisition function $\ddot \rho$ depends only on $\beta$ and $\eta$, and is of the form:
    \begin{equation} \label{eq:naive-rho}
        \ddot \rho = \frac{\lambda^-}{\lambda^-+\lambda^+} = \frac{\eta\beta}{\eta\beta + 1-\beta}
    \end{equation}
\end{proposition}


\subsection{Exact Acquisition Function}

The model above is unrealistic, because it has a positive probability for $\ddot U_{\ddot S} > \beta P$ or $\ddot V_{ \ddot S} > (1-\beta) P$. As the availability of the resource increases, i.e., for larger values of $\alpha$, this probability is non-negligible as subpopulations may saturate. We thus need an exact characterization of $\rho$ that reflects this. Let the disadvantaged and advantaged processes in this case be $U_t$ and $V_t$ respectively. These are now \emph{stopped} versions of the Poisson processes, $\ddot U_t$ and $\ddot V_t$, of the na\"ive case. More precisely, if we define saturation times of these processes as $T_U = \min\{ t ~:~ \ddot U_t \geq \beta P\}$ and $T_V = \min\{ t ~:~ \ddot V_t \geq (1-\beta)P\}$ respectively, i.e., the times when they hit their respective population sizes, then
\[
 U_t=   \left\{ \begin{array}{lcl}
        \ddot U_t &;& \textrm{if } t \leq T_U  \\
        \beta P\phantom{(1-)} &;& \textrm{if } t > T_U
 \end{array}
 \right.
\qquad \qquad
V_t =   \left\{ \begin{array}{lcl}
        \ddot V_t &;& \textrm{if } t \leq T_V  \\
        (1-\beta) P &;& \textrm{if } t > T_V
 \end{array}
 \right.
\]
The exhaustion time is now modified to be: $S=\min \{ t~:~ U_t+V_t\geq N\}$, which is well-defined assuming that no wasteful allocations are allowed, so that $N \leq P$ always. The exact characterization of the acquisition function is now $\rho \define \E{U_{S}} / N$. We have (the proof is given in the supplementary material):

\begin{proposition} \label{prop:exact-rho}
    When both saturation and exhaustion occur, the exact acquisition function $\rho$ depends not only on $\beta$ and $\eta$, but also on $N$ and $P$. It can be written in terms of the na\"ive acquisition function $\ddot \rho$ of Eq. \eqref{eq:naive-rho} as follows:
        \begin{eqnarray} \label{eq:exact-rho}
            \rho  = \ddot\rho \quad
                    &-& \ddot \rho \bar{F}\left(\beta P-1; N-1, \ddot\rho\right)
                    + \frac{\beta P}{N} \bar{F}\left(\beta P; N, \ddot\rho\right) \nonumber \\
                    &+& (1-\ddot \rho) \bar{F}\left((1-\beta)P-1; N-1, 1-\ddot\rho\right)
                    - \frac{(1-\beta) P}{N} \bar{F}\left((1-\beta)P,1-\ddot\rho\right)
    \end{eqnarray}
\end{proposition}

\subsubsection{Tractable Approximation} \label{sec:approx-rho}

The acquisition function directly affects the allocation objective expressed in Eq. \ref{eq:abstract-objective} through the rate disparity, as seen in Eq. \ref{eq:disparity}. While the characterization of $\rho$ in Prop. \ref{prop:exact-rho} is exact, it is not directly amenable to reduction to a convex optimization problem. Fortunately, we can develop a tractable approximation to $\rho$ that also helps develop intuition about its behavior. One can think of the acquisition process in two stages: (1) both subpopulations acquire simultaneously, (2) one subpopulation saturates, while the other continues to acquire.

To quantify these stages, first note that with unlimited resources, the advantaged saturation time is $t=1$, on average. This is because the expected $m ^{\textrm{th}}$ arrival time of $\ddot V$ is $m/\lambda^+$, with $m=\lambda^+=(1-\beta)P$. Similarly, the disadvantaged saturation time is $t=1/\eta$, on average. As a result, it is highly unlikely for the disadvantaged to saturate prior to the advantaged. This sets the stage for the approximation, by taking these times to be equal to their expectations. The strength of such an approximation stems from the fact that for large parameters, a Poisson random variable concentrates very sharply around its mean. The approximate model is:

\begin{itemize}
    \item Before $t=1$, $\E{U_t}$ and $\E{V_t}$ grow linearly at the rate of $\eta\beta P$ and $(1-\beta)P$ respectively. This stage thus reaches a maximal acquisition of $(\eta\beta + 1-\beta)P$.
    \item After $t=1$, $\E{V_t}$ saturates at $(1-\beta)P$, while $\E{U_t}$ continues to grow linearly at the rate of $\eta\beta P$.
    \item Exhaustion happens when $\E{U_t}+\E{V_t}=N$.
\end{itemize}
\vspace{4pt}
In this model, if $N<(\eta\beta + 1-\beta)P$ we have $S\approx (N/P)/(\eta\beta + 1-\beta)$ and thus $\rho = \E{U_S}/N \approx (\eta\beta P S)/N = \ddot \rho$. In other words, if exhaustion occurs during stage (1), the na\"ive model's acquisition function applies. Otherwise, the acquisition function is fully determined by the size of the saturated subpopulation, since all that is left for the other is the remainder of the resources. More precisely, in that case we have $\rho \approx [N-(1-\beta)P]/N$.  We can write:

\[ 
    \rho \approx \left\{ \begin{array}{lcl}
    \frac{\eta \beta}{\eta\beta + 1-\beta} &;& \textrm{if } N<\frac{1}{\alpha} P (\eta\beta + 1 - \beta) \\
    1-\frac{(1-\beta)P}{N} &;& \textrm{otherwise}.
    \end{array} \right.
\]

Further, using the fact that $N/P=\alpha n/p$ and observing that $\frac{\eta \beta}{\eta\beta + 1-\beta} > 1-\frac{(1-\beta)p}{\alpha n} $ is equivalent to $n<\frac{1}{\alpha} p (\eta\beta + 1 - \beta)$, we can write the approximate acquisition function as:
\begin{equation}\label{eq:approx-rho}
    \tilde \rho \define \max \left\{\ddot\rho, 1-\frac{(1-\beta)p}{\alpha n}\right\}
\end{equation}

\begin{figure}[hb]
    \centering
    \subcaptionbox{Acquisition Behavior \label{fig:exact-vs-naive}}
    {\includegraphics[width=0.3\textwidth]{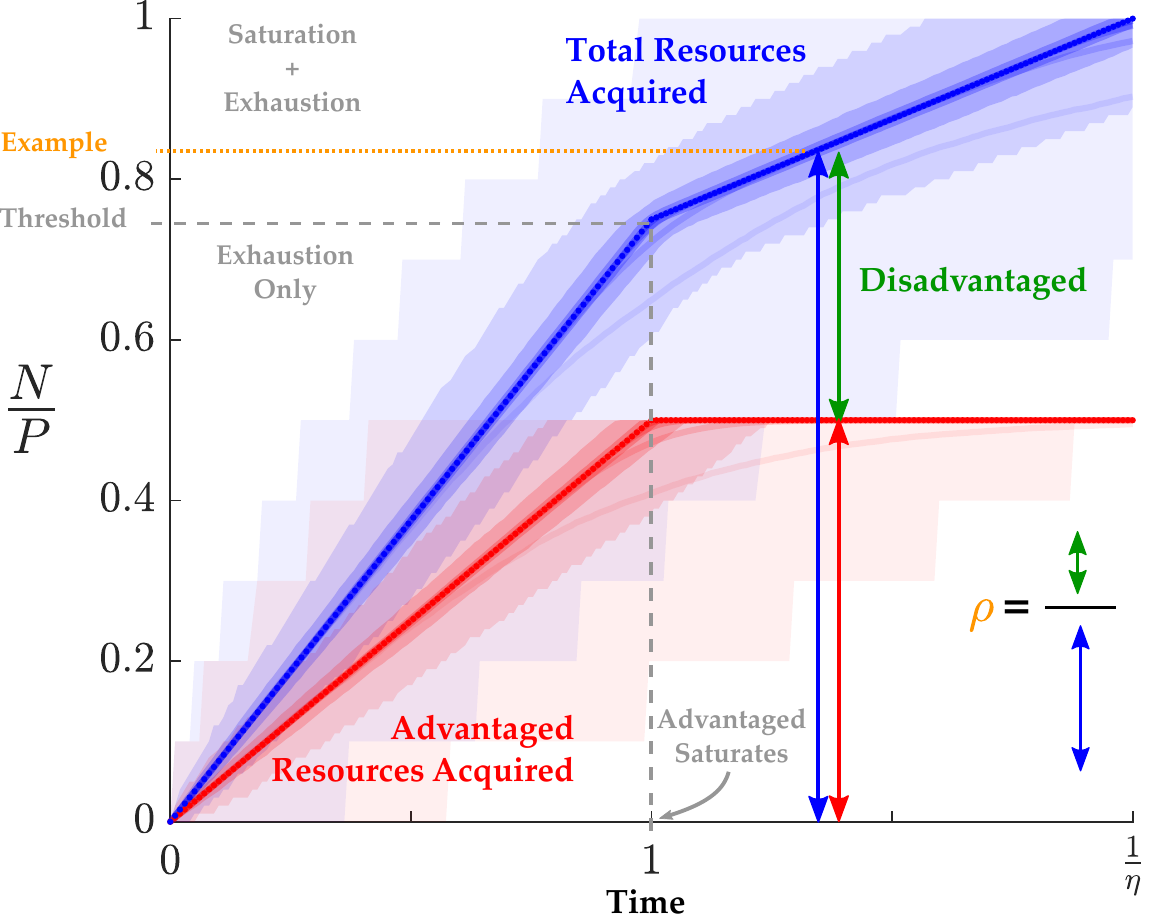}}
    \hspace{0.25\textwidth}
    \subcaptionbox{Robustness to $\eta$ \label{fig:robustness}}
    {\includegraphics[width=0.3\textwidth]{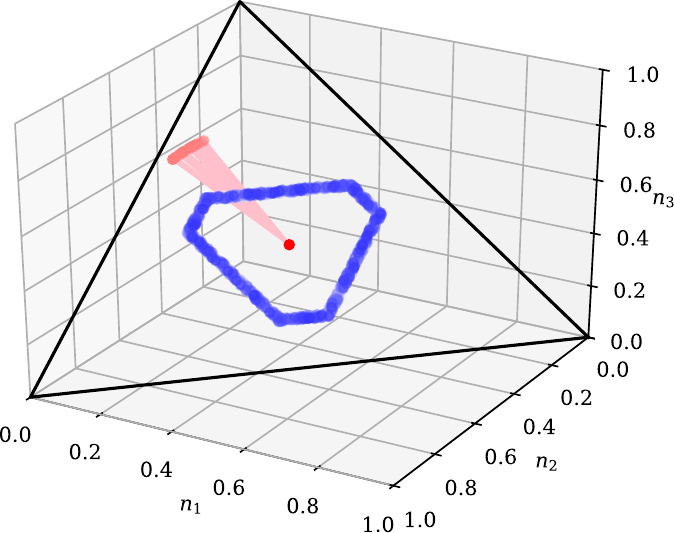}}
    
    \caption{\textbf{Visualizations} --- (\subref{fig:exact-vs-naive}) Acquisition behavior versus time (with $\eta=0.5$, $\beta=0.5$). Blue and red plots are resources acquired by all and by the advantaged, respectively for $P=10, 100, 1000, 10000$ (faintest to darkest), for the maximal expected time. For a specific $N$, resources are exhausted sooner (the orange example). For each case, the average resource is the solid line. The error bars represent the $5$ to $95$ percentile range. The limits (dotted red and blue) correspond to the approximation model in Section \ref{sec:approx-rho}. These are indistinguishable from the average $P=1000$ case and from the entire confidence range for $P=10000$. The acquisition function is always the expected fraction of resources acquired by the disadvantaged upon exhaustion. This can have two distinct behaviors, based on whether only exhaustion occurs or both saturation and exhaustion occur. (\subref{fig:robustness}) Constraint set of the example in Sec. \ref{sec:robustness} (interior of the blue region). Direction of the vector $c_j$ of $\RD$ in Eq. \eqref{eq:disparity-alt}, as $\eta$ varies (in pink). The narrow fan implies that there exists a single discrepancy-optimizing allocation that is optimal for all $\eta$, and thus robust to lack of specification of $\eta$. \label{fig:visualization}}
    
\end{figure}

\looseness=-1 Eq. \eqref{eq:approx-rho} clearly demarcates the two possible behaviors of $\tilde \rho$: the first dominated by exhaustion ($\ddot \rho$) and the second dominated by saturation. Note that the absolute sizes of $N$ and $P$ do not affect this approximation, only their relative size $N/P$ matters. In fact, we can think of this approximation as the limit of $\rho$ for large $N$ and $P$. Fig. \ref{fig:exact-vs-naive} illustrates this visually.

\section{Resource Disparity and Downstream Impact} \label{sec:impact}

\looseness=-1Reducing resource disparity may be thought of as a goal in its own right, as a form of virtue or deontological ethics. However, in the present context one may argue that this goes hand in hand with a more consequentialist notion. Assume that, in the absence of the resource, the advantaged are likely to incur an adverse outcome with probability $x$ and that the disadvantage are more likely to incur such an outcome, say with probability $(1+\delta)x$. These outcomes represent an abstraction of potential downstream impact. In the presence of the resource, say the adverse outcome's likelihood drops by a factor $q\in(0,1)$ for the advantaged and by a factor of $q'$ for the disadvantaged. Thus, with the resource, the advantaged and disadvantaged have a probability $qx$ and $q'(1+\delta)x$ of adverse outcomes respectively. We can then make the following observation.

\begin{proposition} \label{prop:impact} In the adverse outcome scenario described above, the expected number of adverse outcomes is a linear function of $\RD$. Furthermore, the slope of the relationship is positive whenever $\delta > \frac{q'-q}{1-q'}$.
\end{proposition}

\looseness=-1 One can see this by noting that $\RD$ in Eq. \eqref{eq:disparity} is a linear function of the number of vaccines that go to the disadvantaged and that the same is also true for the expected number of adverse outcomes. The condition of Prop. \ref{prop:impact} then follows from a direct calculation. We omit the proof, as it is straightforward. Thus, if the disadvantage is large enough at first (large enough $\delta$) and if the resource allows a decent improvement for the disadvantaged ($q'$ not much larger than $q$) then reducing $\RD$ also reduces the expected number of adverse outcomes. The first facet is intuitive. To see why the second is also necessary, imagine that the resource did nothing to the disadvantaged, i.e., $q' = 1$. In that case, it doesn't help to channel more resources to them, and it is in fact more advantageous (for downstream impact) to channel more to the advantaged. However, these  are arguably satisfied in practice, especially if the resource is even more effective for the disadvantaged ($q'<q$).

\paragraph{Example} Using the CDC estimates \cite{hospitalizationtablescdc} in the period 10/2021 to 2/2022, the average rate of COVID19 hospitalization per 100,000 people in the unvaccinated (advantaged) 18-64 age group is $99$ and the (disadvantage) 65+ group is $415$. For the fully vaccinated, the respective rates are $18$ and $78$. This gives $q=18/99\approx 0.182$ and $q'=78/415\approx 0.188$, corresponding to a minuscule threshold of $0.007$ that $\delta=415/99-1\approx3.19$ readily exceeds. If one considers the 50+ group as advantaged instead, then one gets $q'<q$ and the condition holds trivially. $\square$

\section{Access-Aware Allocation} \label{sec:allocation}

We are now in position to instantiate concrete version of Eq. \eqref{eq:abstract-objective}. To develop intuition, we first do this for the na\"ive acquisition model and show that it reduces to a linear program. We then show how the resulting solution can be used iteratively, to heuristically solve allocation under the approximate access model. As for the exact access model, one could in principle use nonlinear optimization with Eq. \ref{eq:exact-rho} directly, but we do not address this currently. Before proceeding, let us rewrite resource disparity in Eq. \eqref{eq:disparity} as follows:
\begin{equation}
    \RD = \alpha \frac{\sum_j (1-\rho_j) n_j}{\sum_j (1-\beta_j) p_j} - \alpha \frac{\sum_j \rho_j n_j}{\sum_j \beta_j p_j}
    = \sum_j \underbrace{\alpha \left(\frac{1-\rho_j}{\sum_{j'} (1-\beta_{j'}) p_{j'}} - \frac{\rho_j }{\sum_{j'} \beta_{j'} p_{j'}}\right)}_{\textstyle c_j} n_j. \label{eq:disparity-alt}
\end{equation}

\subsection{Na\"ive Allocation} \label{sec:naive-allocation}

In the na\"ive model, resource disparity becomes a linear function of $\mathbf{n}$, because plugging in $\ddot\rho$ instead of $\rho$ in the expression for $c_j$ in Eq. \eqref{eq:disparity-alt} gives:
\[
    c_j = \frac{\alpha}{\eta \beta_j + 1-\beta_j}\left(\frac{1-\beta_j}{\sum_{j'} (1-\beta_{j'}) p_{j'}} - \frac{\eta\beta }{\sum_{j'} \beta_{j'} p_{j'}}\right),
\]
which does not depend on $\mathbf{n}$. Furthermore, the no-waste constraint is already linear and the constraints with both $\ellone(\mathbf{n},\mathbf{p})$ and $\ellrelinf(\mathbf{n},\mathbf{p})$ of Eq. \eqref{eq:distances} can be expanded into a set of linear constraints, via slack variables. As such, in this case the allocation problem \eqref{eq:abstract-objective} can be written as a linear program. Here are the instances that result for each distance:

\begin{equation}
 \begin{array}{lcrclcll}
    \mathbf{n}^\star_{\textcolor{red}{1}}
    &=& \!\!\!\!\!\!\argmin             && \sum_j c_j n_j &&  &\\
    & & \!\!\!\!\!\!\textrm{subject to} &&     n_j,s_j &\geq& 0 & \textrm{\multirow{4}*{$\left.\rule{0em}{10mm}\right\}\, \forall j$}}\\
    & &                     &&     n_j-s_j &\leq& p_j &\\
    & &                     &&    -n_j-s_j &\leq& -p_j &\\
    & &                     &&  \alpha n_j &\leq& p_j &\\
    & &                     &&  \textcolor{red}{\sum_j s_j} &\textcolor{red}{\leq}& \textcolor{red}{\varepsilon} &\\
    & &                     &&  \sum_j n_j &=& 1&
    \end{array}
 \begin{array}{lcrclcll}
    \textrm{or}\quad \mathbf{n}^\star_{\textcolor{red}{\infty}}
    &=& \!\!\!\!\!\!\argmin             && \sum_j c_j n_j &&  &\\
    & & \!\!\!\!\!\!\textrm{subject to} &&     n_j,s_j &\geq& 0 & \textrm{\multirow{5}*{$\left.\rule{0em}{11mm}\right\}\, \forall j$}}\\
    & &                     &&     n_j-s_j &\leq& p_j &\\
    & &                     &&    -n_j-s_j &\leq& -p_j &\\
    & &                     &&  \alpha n_j &\leq& p_j &\\
    & &                     &&  \textcolor{red}{s_j} &\textcolor{red}{\leq}& \textcolor{red}{\varepsilon p_j} &\\
    & &                     &&  \sum_j n_j &=& 1&.
    \end{array} \label{eq:linear-programs}
\end{equation}


\subsection{Better Allocation} \label{sec:better-allocation}

Eq. \eqref{eq:linear-programs} demonstrates concretely how the access model influences the allocation decision. Access determines resource disparity, which in turn can be traded off with geographic disparity. While conveniently in linear program form, these do not capture the phenomenon of saturation that will occur whenever sufficient resources are available at a given location. The exact acquisition function in Eq. \eqref{eq:exact-rho} does address this, but is highly nonlinear. Let us consider instead the approximate acquisition function in Eq. \eqref{eq:approx-rho}. The resulting $\RD$ is still nonlinear, because $\tilde\rho_j$ depends on $n_j$ and $p_j$, and thus so will $c_j$. Moreover, at each location, Eq. \eqref{eq:disparity-alt} multiplies Eq. \eqref{eq:approx-rho} by $-n$, yielding the minimum of two linear functions, which is thus concave. Since $\RD$ is the sum of these functions, the result is a concave minimization problem, which can generally be non-tractable. That said, this informs us that the solution remains on the boundary of the constraint set. In addition, the two distinct behaviors of $\tilde\rho$ suggest that one could start with a na\"ive model and then ``correct'' the model and repeat. This gives the following iterative heuristic:
\begin{enumerate}
    \item Solve the na\"ive allocation problem \eqref{eq:linear-programs}, assuming that exhaustion does not occur.
    \item Use the current allocation to detect exhaustion locations $j$ and correct $\tilde \rho$ according to \eqref{eq:approx-rho} at these locations.
    \item Re-solve the na\"ive allocation problem \eqref{eq:linear-programs}, but now with the $c_j$ weights in \eqref{eq:disparity-alt} using the updated $\tilde\rho_j$.
    \item Repeat Steps (2) and (3) until convergence.
\end{enumerate}
\looseness=-1 This is a heuristic both because $\tilde\rho$ is an approximation, even if an accurate one, and also because a priori the iterations could either not converge or converge to a suboptimal stable point. Nevertheless, in practice the iterations converge extremely quickly and to good solutions, i.e., solutions that significantly reduce resource disparity, as we illustrate experimentally in Sec. \ref{sec:empiricalval}. In the supplementary material, we empirically verify the quality of these using exhaustive search.


\subsection{Robustness to the Access Gap} \label{sec:robustness}

To consider the allocation scheme of Sec. \ref{sec:better-allocation} as a viable alternative to the baseline proportional allocation, one should be able to readily implement it. Apart from design choices, such as the type of allowable deviation from proportionality, all variables are known but one: the access gap $\eta$. How should this be addressed? One approach is to fit the model to historical data, e.g., using a similar approach to the one we use to provide empirical evidence for the access model in Sec. \ref{sec:access-evidence}. While viable, this requires careful elimination of confounding variables. Surprisingly, we experimentally observe that $n_1^\star$ and $n_\infty^\star$ are \emph{robust} to the choice of $\eta$, in that the solution remains stable for a large range of values and in some cases, for \emph{all} $\eta$. We give here an intuitive understanding of this robustness, through a simple example. In the supplementary material, we suggest more direct means to address not knowing $\eta$.

\paragraph{Example} \looseness=-1 Consider a case with only three locations, $j=1,2,3$. Let the population be evenly distributed, $\mathbf{p}=\left(\tfrac{1}{3},\tfrac{1}{3},\tfrac{1}{3}\right)$. Say locations have varying proportions of disadvantaged: low $\beta_1=0.2$, medium $\beta_2=0.5$, and high $\beta_3=0.8$. Assume resources are limited to $70\%$ of the population, $\alpha=0.7$. We are willing to deviate our allocation as much as $\varepsilon=0.4$ away from $\mathbf{n}$ in $\ell_1$, to reduce the resource disparity $\RD$ between the disadvantaged and advantaged. Fig. \ref{fig:robustness} displays the constraint set centered around the interior point $\mathbf{p}$. The fan shape represents the direction of the objective function vector $c_j$ in $\RD$ of Eq. \eqref{eq:disparity-alt} at the optimal allocation, as $\eta$ varies. Note that $\eta$ does not affect the constraint set. Even if $\RD$ depends nonlinearly on $\mathbf{n}$, this informs us that the optimal allocation is indeed at the upper left corner $n^\star_1=\left(\tfrac{2}{15},\tfrac{41}{105},\tfrac{10}{21}\right)$ of the constraint polytope, since otherwise an infinitesimal gradient descent would move away. This implies that the solution is completely unaffected by $\eta$! Interestingly, this phenomenon also occurs with most of the real data cases that we cover in Sec. \ref{sec:behavior-outcome}. Pragmatically, this means that we have an alternative to proportional allocation that \emph{does not require knowledge of the access gaps} and yet results in a much more equitable distribution of resources. Note that the resulting allocation is intuitive: it boosts the highly disadvantaged location's allocation, reduces the low's allocation, and keeps the medium's roughly the same. $\square$


\section{Empirical Validation} \label{sec:empiricalval}

\subsection{Evidence for the Access Model} \label{sec:access-evidence}

To determine whether the proposed model of access has merit in practice, we look at how COVID-19 vaccine acquisition varies across locations. We cannot directly observe the number of vaccines acquired by the advantaged. We can, however, observe the overall vaccination rate. According to the the na\"ive model, i.e., assuming no saturation for simplicity, at any time $t$ the overall vaccination rate is $\E{\ddot U_t + \ddot V_t}/(Pt) = (\eta\beta + 1-\beta) = -(1-\eta)\beta + 1$. In what follows, we empirically show that we indeed observe this linear dependence on $\beta$, despite the potential presence of confounding variables.

\paragraph{Data} For the rates of COVID-19 vaccines, we use publicly available timeseries data from the CDC at the county level \cite{CDCTime}. This data is provided to the CDC by state or territory health departments by healthcare providers. A few counties have no reported vaccinations, we discard these. The features we captured were the dates (\texttt{Date}), county name (\texttt{Recip\_County}), state name (\texttt{Recip\_State}), number of people who received 1 dose of a COVID-19 vaccine (\texttt{Administered\_Dose1\_Recip}), and their percentage at the county level (\texttt{Administered\_Dose1\_Pop\_Pct}). This data reflects recipients over the age of 5. 
To characterize each county's overall vulnerability, we use publicly available data from Surgo Ventures' Precision for Community Vulnerability to COVID-19: Explore the U.S. Data Tool \cite{surgo}. They look at many of the factors the CDC uses to judge vulnerability for COVID-19 such as socioeconomic status, household composition, racial status, existence of pre-exiting conditions, and accessibility to healthcare and transportation. They use these factors to calculate overall vulnerability at the county-level. From Surgo, we captured county name, state name, and vulnerability at $(\textsf{very high}, \textsf{high}, \textsf{moderate}, \textsf{low}, \textsf{very low})$ levels. To explore the applicability of our observations at every scale, we also look at the whole world, using Gallup numbers for vulnerability \cite{gallupdata} and vaccination numbers from One World in Data \cite{mathieu2021global}. We include only a subset of $142$ countries that have clean representation across data sets. \looseness=-1

\begin{figure}[t]
    \centering
    \subcaptionbox{U.S. States\label{fig:access-evidence}}
    {\includegraphics[width=0.5\textwidth]{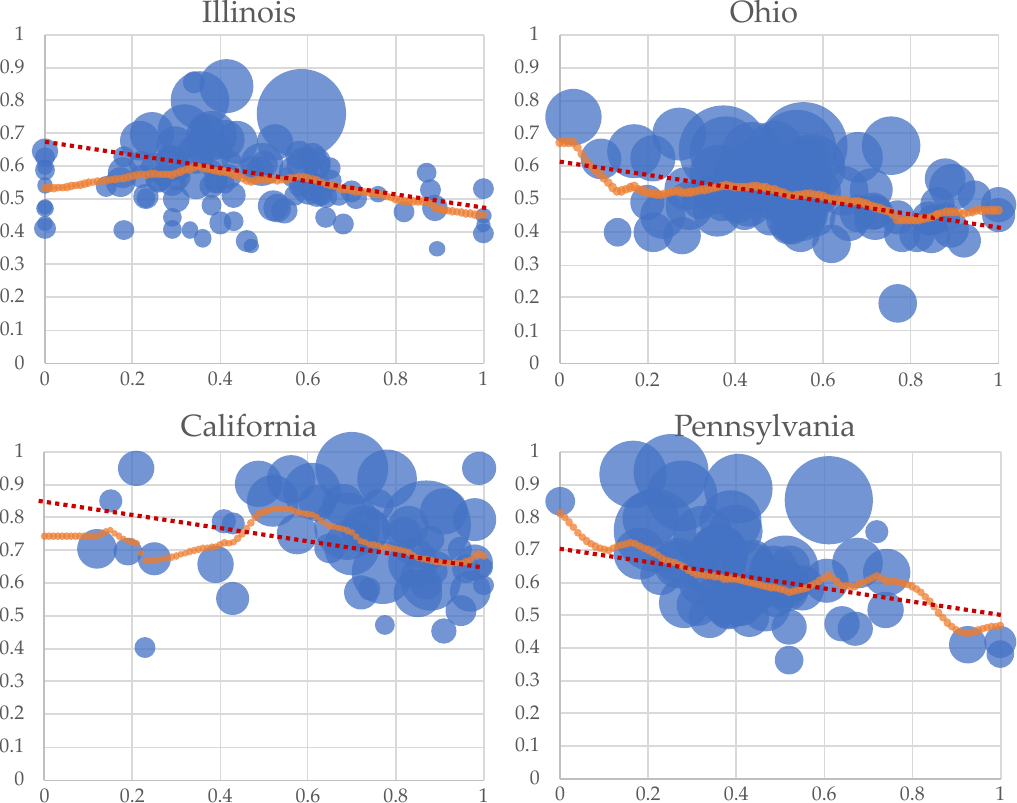}}
    \hspace{0.08\textwidth}
    \subcaptionbox{Global\label{fig:wholeworld}}
    {\vspace{6pt}\includegraphics[width=0.39\textwidth]{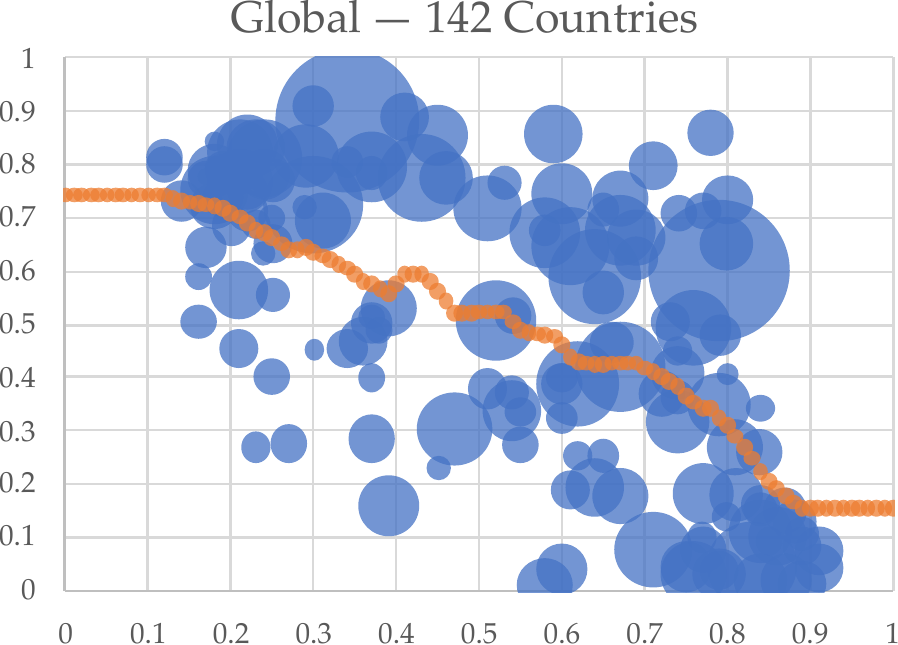} \vspace{20pt}}
    
    \caption{\textbf{Vaccination rate vs. vulnerability} --- (\subref{fig:access-evidence}) Each bubble represents a county, with area proportional to its population. The $x$-axis is the estimated percentage of vulnerable in the county ($\beta$). The $y$-axis is the percentage of the overall population with their first COVID-19 vaccination dose as of 12/30/2021. The orange curve is a soft nearest-neighbor interpolation $\hat y(\beta)$ of the relationship between $\beta$ and $y$. The dotted red is a visual guide illustrating the common dominant behavior across states. (\subref{fig:wholeworld}) Global manifestation of the same phenomenon as in Fig. \ref{fig:access-evidence} --- A roughly linear inverse relationship between vulnerability numbers and vaccination rates across countries.\looseness=-1
    }    

\end{figure}

\paragraph{Experiments} For each county in California, Illinois, Ohio, and Pennsylvania, we look at the percentage of the population that has acquired at least the first-dose of the COVID-19 vaccine on December 30, 2021. We do the same at the global level, per country. Let's call these numbers $y_j$. At the state level, we consider the proportion of the vulnerable at each county to be $\beta_j = \textsf{moderate}_j + \textsf{high}_j + 0.5 \times \textsf{very high}$, where the very high vulnerability range is slightly weighted down because it is often noisy and incomplete. For each of the 4 states, Fig. \ref{fig:access-evidence} shows a scatter plot of $y$ (the bubbles have areas proportional to population size), along with its soft nearest-neighbor interpolation (with $\lambda=20$): 
$
    \hat y(\beta) = \frac{\sum_j y_j \exp(-\lambda |\beta-\beta_j|)}{\sum_j \exp(-\lambda |\beta-\beta_j|)}.
$
In the global case, we set  $\beta_j = \textsf{moderate}_j + \textsf{high}_j$, the 2 categories of Gallup, and illustrate the parallel results in Fig. \ref{fig:wholeworld}.

\paragraph{Results} We observe that, despite noise and confounding variables (e.g., differences between urban/suburban/rural regions, the deliberately ignored saturation which affects the low-$\beta$ range, noise and outliers, etc.), the general trend of $y_j$ versus beta is not only monotonically decreasing, but is indeed roughly linear. In particular, we do see clear segments of linear behavior for primary clusters of counties. We conclude that, though highly idealized, the presented access model has the potential to capture a realistic correlation between the disadvantaged's lack of resource and access obstacles. Note also the common behavior across states, highlighted with the red dotted line in Fig. \ref{fig:access-evidence}. We conjecture that this may be formalized as a methodology to learn $\eta$ from the data. More strikingly, the same behavior extends all the way to the global scale, showing the ubiquity of the phenomenon anticipated by our model.



\subsection{Behavior and Outcomes of Access-Aware Allocation} \label{sec:behavior-outcome}

We now explore, numerically, the behavior and outcomes to expect when employing the proposed access-aware allocation, in comparison to proportional allocation. The latter has a simple characterization: we always have $\frac{n_j}{p_j}=\alpha$. To assess how access-aware allocation deviates from proportional allocation, we visualize the behavior of $\frac{n_j}{p_j}$ in this case, for locations with varying $\beta_j$. In order to demonstrate the possible gains achieved versus proportional allocation, we simultaneously visualize the latter's discrepancy $\RD$ along with the discrepancy achieved by access-aware allocation.

\paragraph{Data} In addition to the county-level CDC data, we also used data privately communicated by the Ohio Health Department, including tract-level maps with information about the number of vulnerable people in each tract, for 22 counties in Ohio\cite{deloittemaps}. The criterion for vulnerability in this case differs from Surgo's, but is based on similar metrics. Using tract-level data from the Census Bureau \cite{tractlevelohio}, we could then determine the population size at each tract, and thus the percentage vulnerable ($\beta$) in each. This allows us to explore the applicability of the present methodology at such finer granularity. We also explore a much coarser granularity, with allocation of vaccines across all $50$ states in the US. The data for this is an aggregation of that obtained from CDC and Surgo, as described in Sec. \ref{sec:access-evidence}. The vulnerable percentage in this section is taken to be everyone categorized as $\textsf{very high}$, $\textsf{high}$, or $\textsf{moderate}$.

\paragraph{Experiments} We fix the value of $\epsilon$ to $0.1$ throughout. The influence of $\epsilon$ on the allocation is apparent nonetheless, and we comment on it below. We look at the states of California, Illinois, Georgia, Ohio, and Pennsylvania (locations = counties), in addition to Franklin County, OH (locations = Census tracts),  and the US (locations = states). We consider $\alpha=0.1$ (low), $0.5$ (medium), and $0.9$ (high) levels of resource availability. We vary $\eta$ from $0$ to $1$. In each scenario, we solve for $n_1^\star$ and $n_2^\star$ using the iterative approach in Sec \ref{sec:behavior-outcome}, by solving Eq. \eqref{eq:linear-programs} and alternating with updates of $\rho$.

\begin{figure}[t]
    \centering
    \subcaptionbox{Behavior of $n^\star_1$\label{fig:ellone-behavior}}
    {\includegraphics[width=0.48\textwidth]{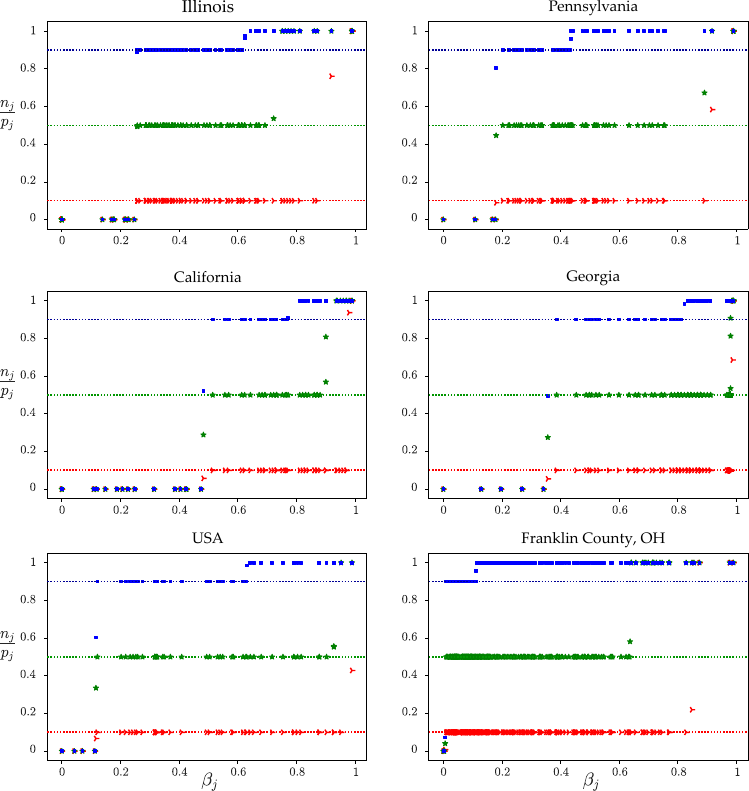}}
    \subcaptionbox{Behavior of $n^\star_\infty$\label{fig:ellrelinf-behavior}}
    {\includegraphics[width=0.48\textwidth]{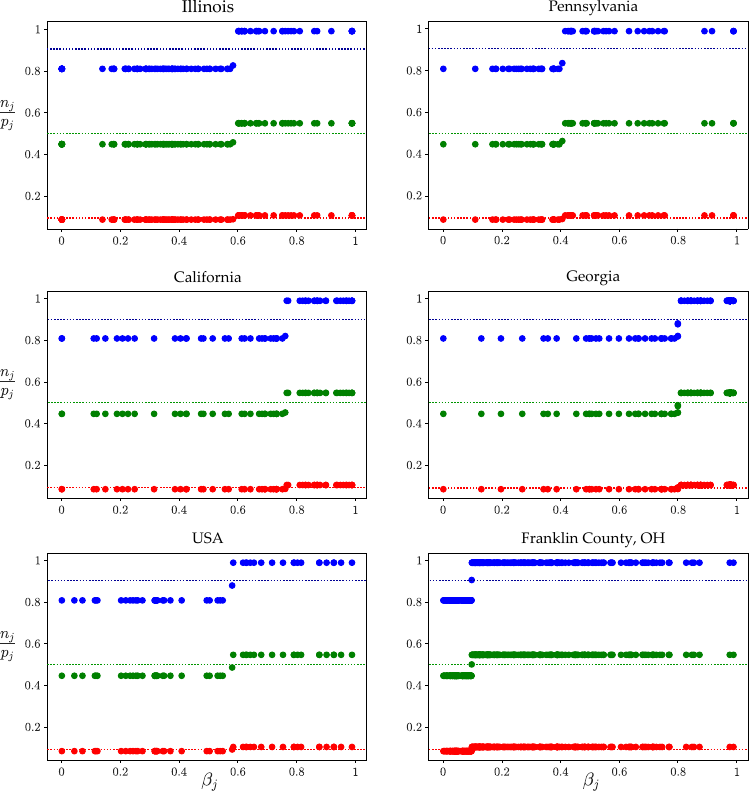}}
    \caption{\textbf{Behavior of Allocations} --- Each point is a location $j$. The $x$-axis is the percentage vulnerable $\beta$ at location $j$ and the $y$-axis is the ratio of allocation to its population $\frac{n}{p}$. The allocation for \textcolor{red}{$\alpha=0.1$}, \textcolor{green}{$\alpha=0.5$}, and \textcolor{blue}{$\alpha=0.9$} are also indicated with dotted lines. Main observations --- (\subref{fig:ellone-behavior}) $n^\star_1$ specifies two thresholds, does not allocate below the lower, fully allocates above the higher, and maintains $\alpha$ in between. (\subref{fig:ellrelinf-behavior}) $n^\star_\infty$ fixes a threshold and alters the proportional allocation slightly lower below it and slightly higher above it.\label{fig:behavior}}
\end{figure}

\paragraph{Results --- Behavior} Fig. \ref{fig:ellone-behavior} illustrates the behavior of the $n_1^\star$ allocation through the ratio $\frac{n_j}{p_j}$. The dotted line corresponds to $\alpha$, the proportional allocation case. We observe that in all cases the access-aware allocation has roughly the same form: no allocation $\frac{n}{p}=0$ below a lower threshold, full allocation $\frac{n}{p}=1$ above an upper threshold, and proportional allocation $\frac{n}{p}=\alpha$ in between. The value of $\epsilon$ influences how close the upper and lower thresholds are for fixed $\alpha$. This is a harsh policy and is due to the lax nature of $\ud_1$ that allows such deviation from proportionality. It is worth noting that this is not always the form of $n_1^\star$, as we have seen in the example of Sec. \ref{sec:robustness}. 

Fig. \ref{fig:ellrelinf-behavior} illustrates the behavior of the $n_\infty^\star$ allocation, which is much simpler: slightly less allocation below a threshold $\frac{n}{p}<\alpha$ and slightly more allocation above a threshold $\frac{n}{p}>\alpha$. The value of $\epsilon$ determines the amount of change from $\alpha$. It is also interesting to note that in this case, unlike the thresholds of $n_1^\star$, this threshold does not depend on $\alpha$. This is a lax policy, giving a much gentler deviation from proportionality, due to the harsh nature of $\ud_\infty$.

Lastly, we note that we do not display multiple policies for each $\eta$, because under each scenario the policy is unaffected by the choice of $\eta$, thus supporting the claims of robustness in Sec. \ref{sec:robustness}.

\begin{figure}[b]
    \centering
    \subcaptionbox{Outcomes of $n^\star_1$\label{fig:ellone-outcome}}
    {\includegraphics[width=0.48\textwidth]{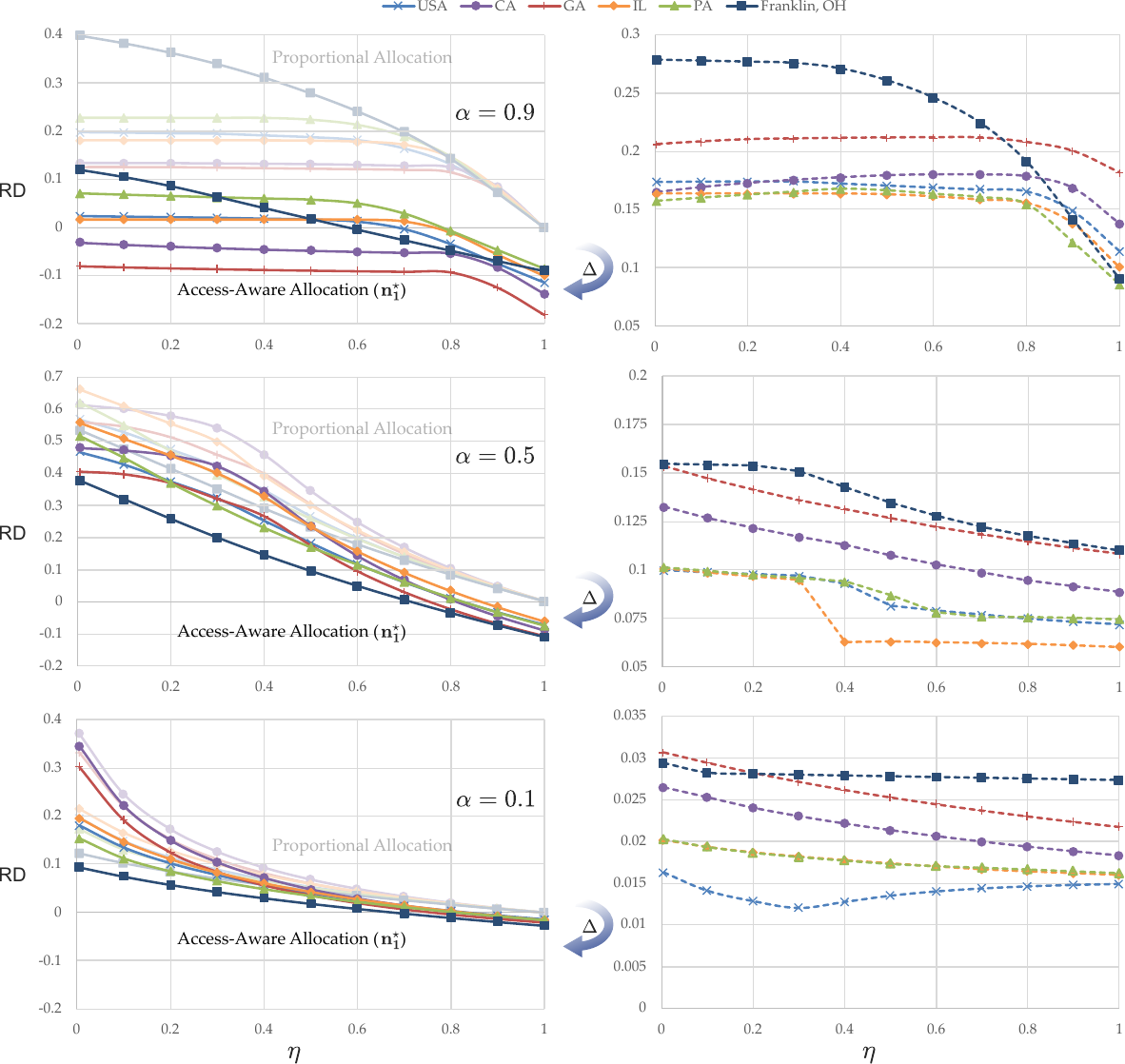}}
    \subcaptionbox{Outcomes of $n^\star_\infty$\label{fig:ellrelinf-outcome}}
    {\includegraphics[width=0.48\textwidth]{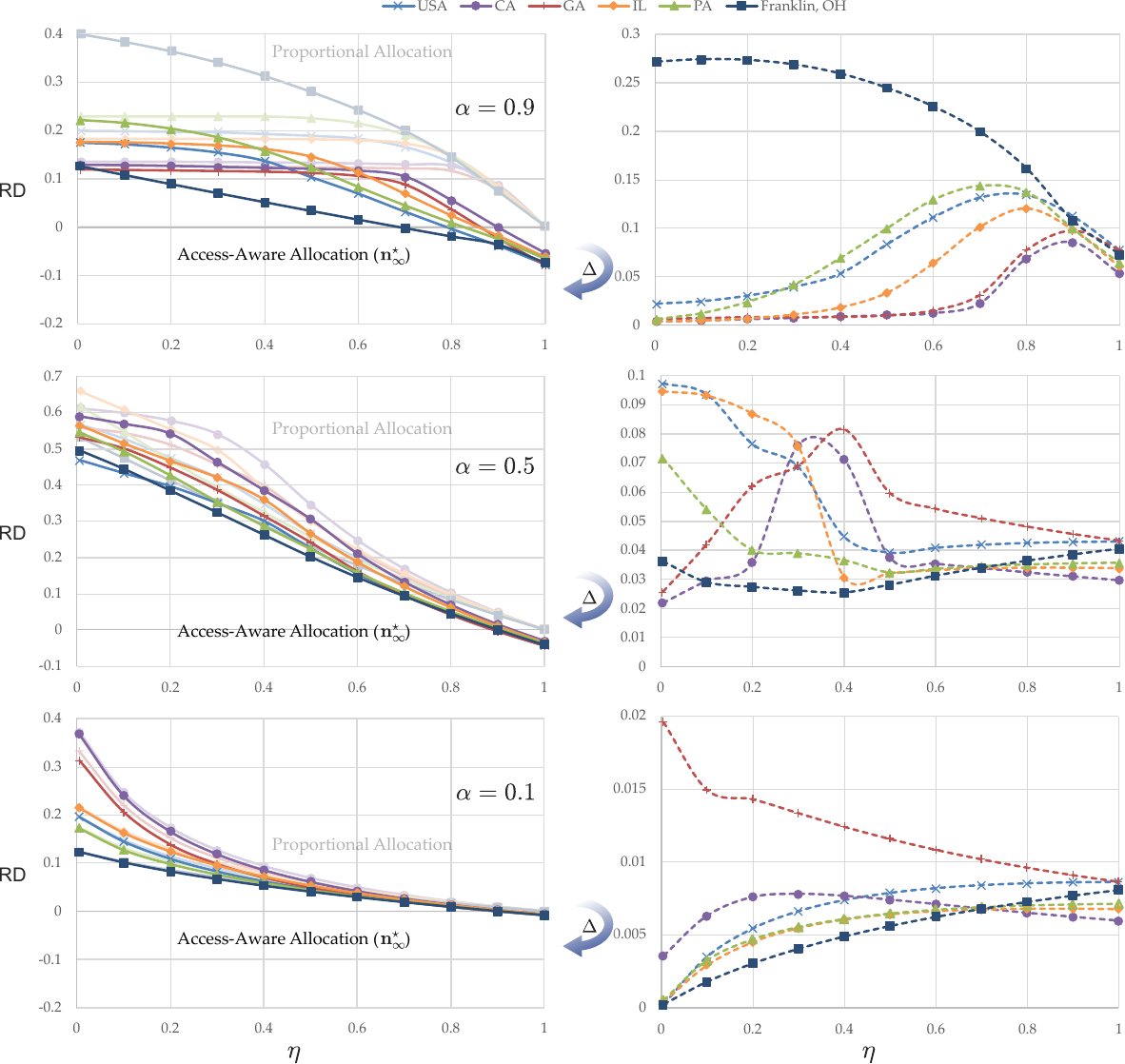}}

    \caption{\textbf{Outcomes of Allocations} --- (Left) Resource (vaccination) disparity ($y$-axis, $\RD$) of both the access-aware allocation $n^\star_\infty$ offers and the proportional allocation (faint plots), for $\alpha=0.1, 0.5, 0.9$. (Right) Difference between the two, which captures the improvement in disparity by using the access-aware allocation. The $x$-axis is the access gap, $\eta$. Main observations --- (\subref{fig:ellone-outcome}) For $n^\star_1$, improvements are more marked when the access gap is larger, i.e. when $\eta$ is smaller. Counterintuitively, more availability (larger $\alpha$) offers relatively more opportunity to mitigate disparity, as explained in the text. (\subref{fig:ellrelinf-outcome}) For $n^\star_\infty$, the general behavior is comparable to that of $n^\star_1$. However, in this case we are more constrained not to deviate from proportionality, and the gains are thus less.\label{fig:outcome}}

\end{figure}

\paragraph{Results --- Outcomes} Fig. \ref{fig:ellone-outcome} and Fig. \ref{fig:ellrelinf-outcome} illustrate the improvement in resource disparity that $n_1^\star$ and $n_\infty^\star$ afford. Each constraint results in specific characteristics. As we could expect from their harsh and lax respective behaviors, $n_1^\star$ can deliver a more marked improvement to resource disparity than $n_\infty^\star$ can.

More interestingly, under $n_1^\star$, access-awareness is generally (though not always) more impactful at higher access gaps (small $\eta$) and less so at lower gaps (large $\eta$). There isn't such a monotonic relationship for $n_\infty^\star$, with $\RD$ often peaking in $\eta$ in that case. This seems to suggest that when too much deviation from proportionality isn't desirable (so one would best use $\ud_\infty$), then allocation, to be effective, needs to be compounded with enough social effort to bring the disadvantaged above a certain level of access beyond which it is able to mitigate any remaining difference. However, if more bold redistribution is accepted (thus one could use $\ud_1$) then allocation on its own can go a long way to mitigate disparity. Thus this work reveals an interesting tradeoff. To borrow some political jargon, if one is ``conservative'' in resource allocation, they ought to be more ``progressive'' in providing social safety nets. Conversely, if one is ``socialist'' in resource allocation, they can afford to be less regulated and more ``libertarian'' when it comes to society.

Both instances do however share a common yet counterintuitive pattern, in that the disparity reduction is relatively higher when resources are more available ($\alpha=0.9$), especially in moderate access gaps (mid to high $\eta$). This is due to saturation effects, as without access-awareness, the advantaged subpopulation can quickly acquire all the needed resources, leaving scraps to the disadvantaged.


What is practically most significant, however, is that at moderate availability ($\alpha=0.5$) the improvement in $\RD$ is consistent in both instances, no matter what the value of $\eta$ is. We see this visually as the $\RD$ plots having shifted down. This highlights the potential impact that access awareness could have on distributing resources more equitably, no matter what the gap is between the advantaged and disadvantaged.

A final phenomenon that is worth clarifying is the following. Imagine $\eta=0$, meaning that no matter how resources are allocated, the disadvantaged acquire \emph{nothing}. In that case, one cannot hope to improve on proportional allocation (or for that matter on any non-wasteful allocation), since all resources will then flow to the advantaged, and no reallocation can change it. One would then expect, near $\eta=0$, for the $\RD$ of the access-aware allocations to approach that of the proportional allocation. Yet, we observe a clear separation between the two. How can we explain this? This is due to the fact that our access model operates with an infinite time horizon, which creates a discontinuity. While the disadvantaged acquire nothing at $\eta=0$, even a modicum of access $\eta>0$ no matter how small, the disadvantaged can acquire whatever is leftover after the advantaged saturate. They can do this, even if it takes them an arbitrarily long time. And this offers an opportunity for access-aware allocation to make a difference. This also opens up an avenue to explore finite time horizon effects by appropriately modifying the access model. 

\section{Discussions} \label{sec:discussions}
Throughout this work, we have considered technical means by which to address issues with access to limited resources for members of vulnerable groups. We acknowledge that this is an oversimplification of society and that there are many factors, both endogenous and exogenous, that can hamper a person's acquisition of a limited resource.

\paragraph{Other Factors of Slowed Access} Often, members of vulnerable groups have been marginalized and mistreated by systems offering aid and assistance to illness or societal issues (e.g. Tuskegee Experiments) \cite{wasserman2007rasing}. This mistreatment has lead to mistrust and reluctance to accepting proposed solutions amongst some subpopulations. In addition, there is often a wave of misinformation distributed in mass on public concerns \cite{naeem2021exploration}. Without the correct information on a resource, people are robbed of the opportunity to decide whether obtaining a resource will truly be beneficial for them. Educating people on the true benefits and disadvantages of obtaining a resource could help in their decision-making. Even with a trustworthy system, true education on obtaining a resource, and the solution of proximity issues, more barriers exist, e.g., lack of internet, inflexible jobs, mobility issues, or no health insurance \cite{jean2021vaccine}.

\paragraph{Calls for Justice} Our approach to mitigating slowed down access through increased proximity could be interpreted as a form of justice, in that those a part of the disadvantaged population have often been marginalized, creating lack of access and exacerbating the effects of this deficit. This marginalization has been caused by multiple factors and reared in various ways, some include: redlining which hurt access to jobs and healthcare \cite{redlining}, social exclusion which can lead to lack of education, and inability to gather those resources \cite{alston2003educational}, etc. Such history has shifted the distribution of power \cite{rediet2021fairness}. Though these effects are deeply rooted and ring throughout the lives of the disadvantaged, the hope is that taking such steps will encourage appropriate visibility of these members of society and, by doing so, work toward reversing injustice. \looseness=-1


\section{Acknowledgements}
This paper is based upon work supported in part by the NSF LSAMP Bridge to the Doctorate Program under Award No. HRD-1906075 and the NSF-Amazon Program on Fairness in AI in Collaboration with Amazon under Award No. IIS-1939743, titled FAI: Addressing the 3D Challenges for Data-Driven Fairness: Deficiency, Dynamics, and Disagreement. Any opinion, findings, and conclusions or recommendations expressed in this paper are those of the authors and do not necessarily reflect the views of the National Science Foundation or Amazon.

\newpage
\bibliographystyle{ACM-Reference-Format}
\bibliography{references}

\newpage

\appendix

\section{Verification of the Heuristic} \label{sec:verification}

\begin{figure}[h]
    \includegraphics[width=0.8\textwidth]{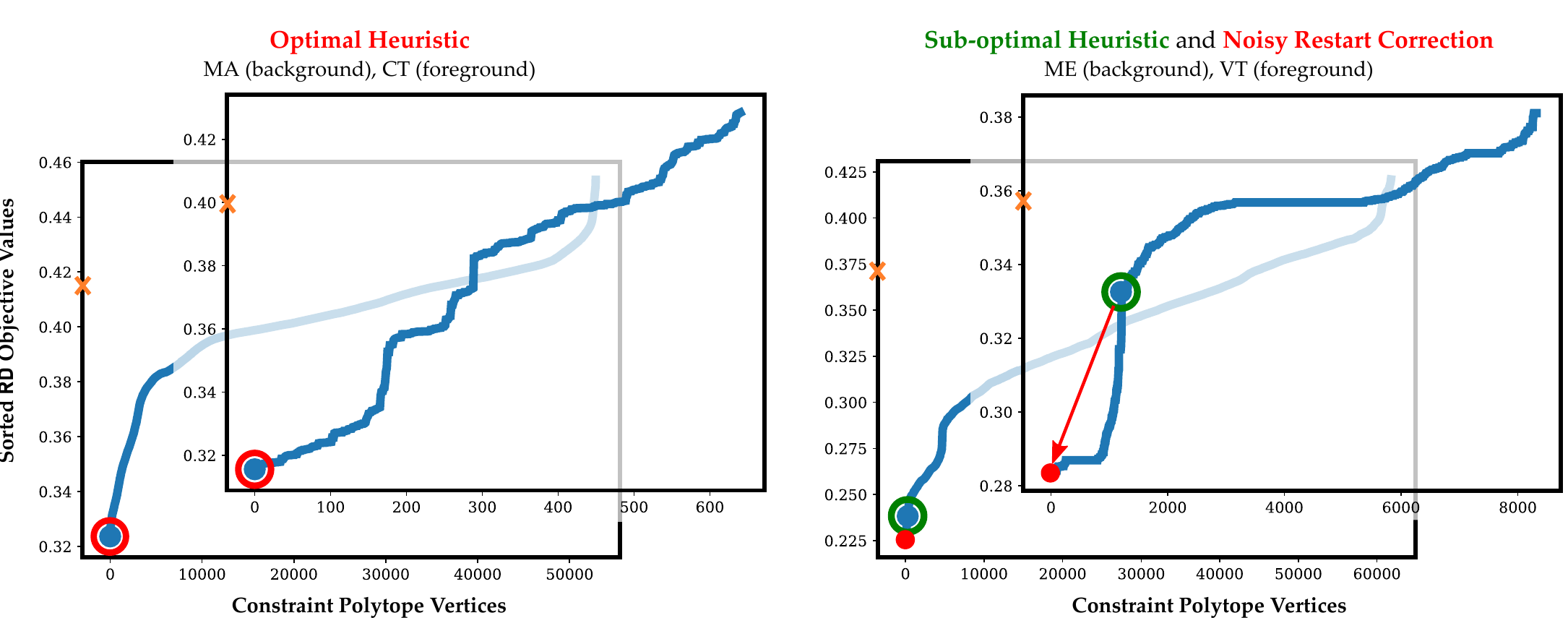}
    \caption{Sorted $\RD$ values at every vertex of the constraint polytope for allocation across counties for $4$ states, CT, MA, ME, and VT. The orange $\times$ indicates the $\RD$ of proportional allocation. The dots represent the $\RD$ of the access-aware allocation, either in the heuristic form of the paper, or augmented via noisy restarts.}
    \label{fig:verification}
\end{figure}

Let us take a closer look at the behavior of the heuristic optimization via successive iteration, as described in Sec. 6.2 and used throughout our experiments. Recall that the underlying problem is a minimization of a piecewise-linear concave objective over a bounded convex polytope. This means that the solution lies on a vertex of this polytope. To empirically verify the heuristic, we can exhaustively test all these vertices. Because such an approach is feasible only at a small scale, we restrict this verification to small state-level examples: states CT, MA, ME, and VT, which have 8, 14, 16, and 14 counties respectively. Vulnerability ($\beta$) is based on moderate to very high percentages from the Surgo data. We fix $\varepsilon=0.1, \alpha=0.5,$ and $\eta=0.3$ as representative values. We employ the Avis--Fukuda algorithm \cite{avis1992pivoting} to enumerate the vertices of the constraint polytope in each case. We then evaluate the objective at each vertex. Fig. \ref{fig:verification} illustrates the sorted objective values across all vertices in each state. In all cases, the heuristic converges. In most cases, it is approximately (ME) or exactly (MA and CT) at the optimal value. However, in some cases it can remain sub-optimal (NH). Note that even then, the access-aware heuristic improves on proportional allocation (the $\times$ mark). We conclude that, awaiting further investigation into better optimization approaches, the access model and the corresponding heuristic allocation presented in this paper are a pragmatic approach to tackle resource disparity. That said, to explore potential avenues to remedy suboptimality, we conjecture that initializing the iterations with a noisy version of the na\"ive acquisition function can dislodge from local minima. Such restarts are much faster than vertex enumeration, which becomes prohibitive with $20+$ locations. In initial experiments, choosing the best out of $100$ random noisy restarts (additive standard normal per coordinate) yielded the optimal solution every time. Completely random restarts performed worse.

\section{No Knowledge of the Access Gap} \label{sec:knowing-gap}

We include her some discussion on potential approaches that could be taken in situations where the methodology of the paper is applied, but where robustness to $\eta$ is not inherent. Let us assume that the range $(0,1)$ of $\eta$ can be partitioned to finitely many, $L$, regions with a representative $\eta_i$ in each (the piecewise linear nature of the problem facilitates this, though does not always guarantee it, cf. the adversarial version in the next paragraph). If a Bayesian estimate of the distribution of the $\eta_i$ is available, then one could also minimize the expected discrepancy $\sum_{i=1}^{L} \prob{\eta=\eta_i} \RD(\mathbf{n},\eta_i)$ with the same methodology as in Sec. \ref{sec:better-allocation}, as it is still a concave function admitting a minimum on the boundary of the convex constraint set. More precisely, the iterations would maintain a separate updated $\tilde\rho_{j}(i)$ and corresponding $c_j(i)$ for each $i$ in Steps (1) and (2), and use the effective $c_j$ to be $\sum_{i=1}^{L} \prob{\eta=\eta_i} c_j(i)$ in Steps (1) and (3).

One may also want to find an adversarially robust allocation via $\min_\mathbf{n} \max_i \RD(\mathbf{n},\eta_i)$. The objective function here can become irregular as a max of concave functions is not even concave anymore, and the the optimal allocation could fall in the interior of the constraint set. In this case, one could relax this minimax problem into its maximin dual, by solving $\max_{\Psymb} \min_{\mathbf{n}} \sum_{i=1}^{L} \prob{\eta=\eta_i} \RD(\mathbf{n},\eta_i)$ instead. Then, the earlier Bayesian solution can be used along with a numerically differentiated interior point method to optimize over $\Psymb$. We can then guarantee that $\min_\mathbf{n} \max_i \RD(\mathbf{n},\eta_i) \in [1, L] \times \max_{\Psymb} \min_{\mathbf{n}} \sum_{i=1}^{L} \prob{\eta=\eta_i} \RD(\mathbf{n},\eta_i)$.

\section{Proofs} \label{sec:proofs}

\subsubsection*{Proof of Prop. \ref{prop:naive-rho}}

Conditionally on $\ddot S=s$, $\ddot U_{\ddot S}$ and $\ddot V_S$ are equivalent to independent Poisson random variables $X$ and $Y$ with respective rates $\mu\define \lambda^-s$ and $\nu\define \lambda^+s$, conditioned on $X+Y=N$. Note that $X+Y$ itself is then Poisson with parameter $\mu+\nu$. To calculate $\E{\ddot U_{\ddot S}|\ddot S=s}$, we can equivalently calculate:
\begin{eqnarray*}
\E{X|X+Y=N} &=& \sum_{x=0}^N x \prob{X=x | X+Y=N}
= \sum_{x=0}^N x\frac{\prob{X=x, Y=N-x}}{\prob{X+Y=N}}
= \frac{\sum_{x=0}^N x \frac{\mu^x \ue^{-\mu}}{x!} \frac{\nu^{N-x} \ue^{-\nu}}{(N-x)!}}{\frac{(\mu+\nu)^N \ue^{-(\mu+\nu)}}{N!}} \\
&=& \frac{N\mu}{(\mu+\nu)} \sum_{x=1}^{N-1} \tfrac{(N-1)!}{(x-1)!(N-x)!}  \left(\tfrac{\mu }{\mu+\nu}\right)^{x-1} \left(\tfrac{\nu}{\mu+\nu}\right)^{N-x} = \frac{N\mu}{(\mu+\nu)},
\end{eqnarray*}
where we used the binomial identity. By substituting for $\mu$ and $\nu$, we get that $\E{U_{\ddot S}|\ddot S=s}/N=\frac{\lambda^-}{\lambda^++\lambda^-}$ regardless of $s$. Taking iterated expectation over $\ddot S$ completes the proof. \hfill $\square$


\subsubsection*{Proof of Prop. \ref{prop:exact-rho}}

Consider the two events $\mathcal A = \{\ddot U_{\ddot S} > \beta P\}$ (the disadvantaged have saturated, $T_U < \ddot S$) and $\mathcal{B} = \{\ddot V_{\ddot S} > (1-\beta)P\}$ (the advantaged have saturated $T_V < \ddot S$). These are disjoint, since otherwise $\ddot{U}_{\ddot S}+\ddot{V}_{\ddot S} > \ddot{P}$, which contradicts the definition of $\ddot{S}$. Along with the complement $\mathcal{C}=(\prob{\mathcal{A}}\cup\prob{\mathcal{B}})^\mathsf{c}$ of their union, these events thus form a partition of the sample space. In $\prob{\mathcal{C}}$, we have $S=\ddot{S}$ (since no saturation occurs) and $U_S = \ddot U_{\ddot S}$. Now let us write the following two total expectations:
\[
    \E{\ddot U_{\ddot S}} = \prob{\mathcal{A}} \E{\ddot U_{\ddot S} |\mathcal{A}} + \prob{\mathcal{B}} \E{\ddot U_{\ddot S}|\mathcal{B}} + \prob{\mathcal{C}} \underbrace{\E{\ddot U_{\ddot S}|\mathcal{C}}}_{\E{U_S|\mathcal C}}
\]
and
\[
    \E{U_{S}} = \prob{\mathcal{A}} \underbrace{\E{U_{S} |\mathcal{A}}}_{\beta P} + \prob{\mathcal{B}} \underbrace{\E{U_{S}|\mathcal{B}}}_{N-(1-\beta)P} + \prob{\mathcal{C}} \E{U_{S}|\mathcal{C}},
\]
where the latter two expressions are due to the saturation events of $\mathcal A$ and $\mathcal B$.
Subtracting the first from the second:
\begin{equation*}
    \E{U_S} = \E{\ddot U_{\ddot S}} 
    + \prob{\mathcal{A}} \left(\beta P - \E{\ddot U_{\ddot S} |\mathcal{A}}\right)
    + \prob{\mathcal{B}} \left(N-(1-\beta)P - \E{\ddot U_{\ddot S}|\mathcal{B}}\right).
\end{equation*}

This is convenient because we have now expressed the saturated process's expectation completely in terms of the unsaturated process. All the quantities to be found depend only on $\ddot S$, $\ddot U_t$, and $\ddot V_t$. Using a similar approach as in the proof of Prop. \ref{prop:naive-rho} (conditioning on $\ddot S=s$, using the equivalent Poisson random variables $X$ and $Y$, then performing iterated expectation), we get:
\[ 
\begin{array}{rclcrcl}
\prob{\mathcal A} &=& 1 - F(\beta P; N, \ddot \rho) && \E{\ddot U_{\ddot S} | \mathcal A} &=& N\ddot\rho \frac{1-F(\beta P-1; N-1, \ddot \rho)}{\prob{\mathcal A} } \\

\prob{\mathcal B} &=& 1 - F((1-\beta)P; N, 1-\ddot \rho) && \E{\ddot V_{\ddot S}|\mathcal B} &=& N(1-\ddot\rho) \frac{1-F((1-\beta)P-1; N-1, 1-\ddot \rho)}{\prob{\mathcal B} } 

\end{array}
\]
Note that $\E{\ddot U_{\ddot S}|\mathcal B} = N - \E{\ddot V_{\ddot S}|\mathcal B} $, and we have already determined that $\E{\ddot U_{\ddot S}}=N\ddot\rho$. Combining these expressions completes the proof.\hfill $\square$

\end{document}